\documentclass{article}

\PassOptionsToPackage{numbers, compress}{natbib}

\usepackage[preprint]{neurips_2023}

\usepackage[utf8]{inputenc} 
\usepackage[T1]{fontenc}    
\usepackage[breaklinks=true,colorlinks,bookmarks=false]{hyperref}      
\usepackage{url}            
\usepackage{booktabs}       
\usepackage{amsfonts}       
\usepackage{nicefrac}       
\usepackage{microtype}      
\usepackage{xcolor}         

\usepackage{algpseudocode}
\usepackage[linesnumbered, ruled]{algorithm2e}

\usepackage{amsmath}
\usepackage{graphicx}
\usepackage{subfig}
\usepackage{dsfont}
\usepackage{wrapfig}
\usepackage{multirow}
\usepackage{arydshln}
\usepackage{utfsym}

\title{Robustness and Generalizability of Deepfake Detection: A Study with Diffusion Models}

\author{%
  Haixu Song \\
  Tsinghua University \\
  Beijing, China \\
  \texttt{shx22@mails.tsinghua.edu.cn} \\
  \And
  Shiyu Huang \\
  4Paradigm Inc. \\
  Beijing, China \\
  \texttt{huangshiyu@4paradigm.com} \\
  \AND
  Yinpeng Dong \\
  Tsinghua University \\
  Beijing, China \\
  \texttt{dongyinpeng@mail.tsinghua.edu.cn} \\
  \And
  Wei-Wei Tu \\
  4Paradigm Inc. \\
  Beijing, China \\
  \texttt{tuweiwei@4paradigm.com} \\
}

\begin{document}

\maketitle

\begin{abstract}
The rise of deepfake images, especially of well-known personalities, poses a serious threat to the dissemination of authentic information. To tackle this, we present a thorough investigation into how deepfakes are produced and how they can be identified. The cornerstone of our research is a rich collection of artificial celebrity faces, titled DeepFakeFace (DFF). We crafted the DFF dataset using advanced diffusion models and have shared it with the community through online platforms\footnote{Access the code on GitHub: \url{https://github.com/OpenRL-Lab/DeepFakeFace/} and the dataset on HuggingFace: \url{https://huggingface.co/datasets/OpenRL/DeepFakeFace}.}. This data serves as a robust foundation to train and test algorithms designed to spot deepfakes. 
We carried out a thorough review of the DFF dataset and suggest two evaluation methods to gauge the strength and adaptability of deepfake recognition tools. The first method tests whether an algorithm trained on one type of fake images can recognize those produced by other methods. The second evaluates the algorithm's performance with imperfect images, like those that are blurry, of low quality, or compressed. 
Given varied results across deepfake methods and image changes, our findings stress the need for better deepfake detectors. Our DFF dataset and tests aim to boost the development of more effective tools against deepfakes.
\end{abstract}

\section{Introduction}

Deepfake technology has become a significant concern in today's digital landscape~\cite{arjovsky2017wasserstein,karras2019style,goodfellow2020generative}. These advanced computer-generated images, known as deepfakes, can mimic real photos so closely that they often deceive viewers. The biggest worry is when these fake images, particularly of famous individuals, are used wrongly to spread misinformation, influence people's views, or even trick security systems~\cite{li2020celeb}. As it becomes harder for us to spot these images by just looking, it's evident we need better tools to detect them.


\begin{figure}[t]
\begin{center}

\subfloat[Images from IMDB-WIKI dataset~\cite{rothe2015dex}.]{\begin{centering}
\includegraphics[width=0.98\linewidth]{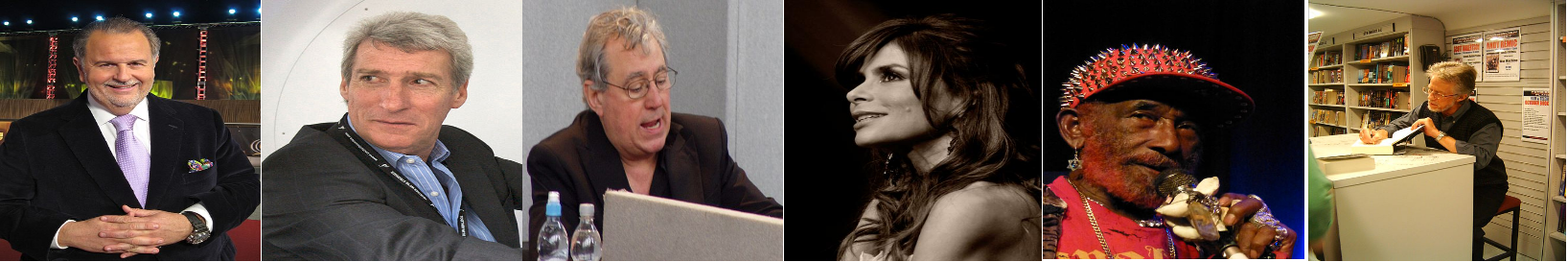}
\end{centering}
}

\subfloat[Deepfake images from FaceForensics++~\cite{rossler2019faceforensics++}.]{\begin{centering}
\includegraphics[width=0.98\linewidth]{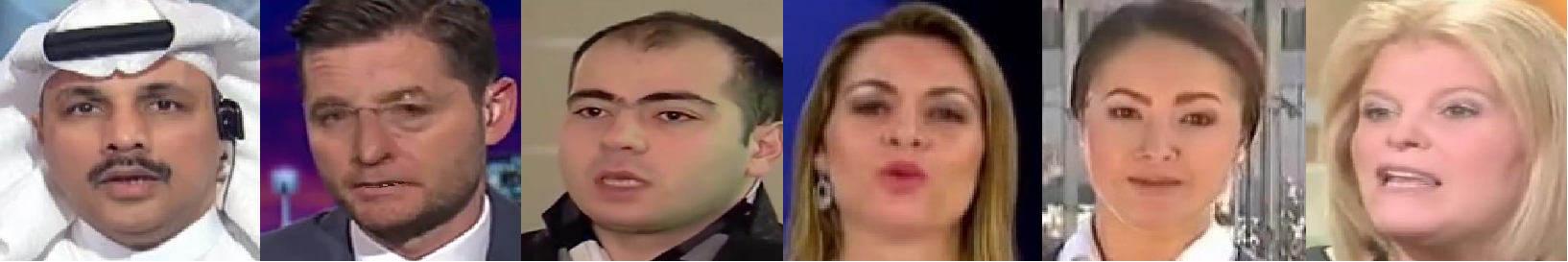}
\end{centering}
}

\subfloat[Deepfake images generated via Stable Diffusion v1.5.]{\begin{centering}
\includegraphics[width=0.98\linewidth]{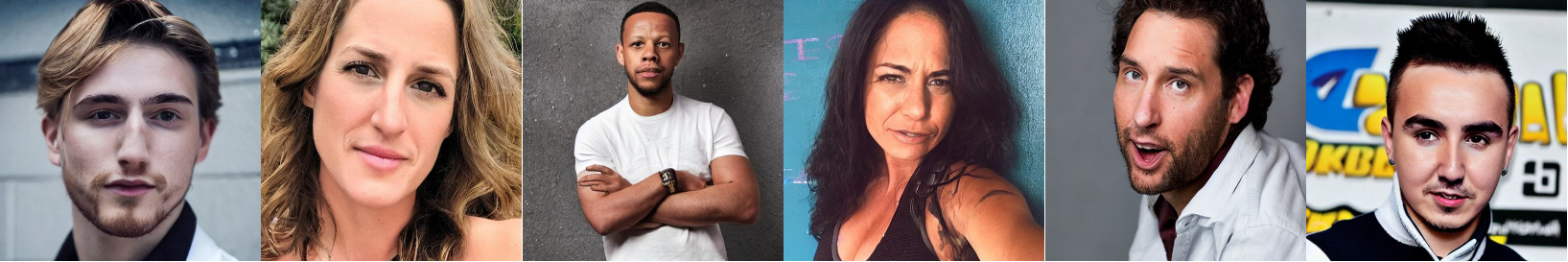}
\end{centering}
}

\subfloat[Deepfake images generated via Stable Diffusion Inpainting.]{\begin{centering}
\includegraphics[width=0.98\linewidth]{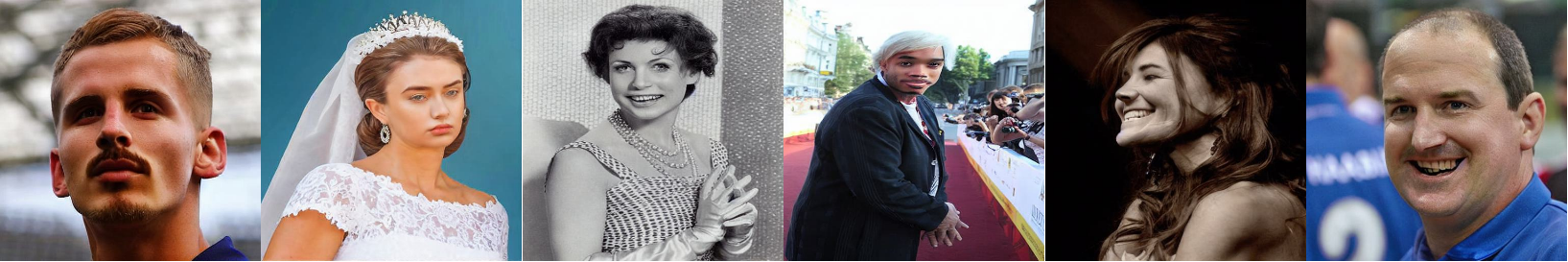}
\end{centering}
}

\subfloat[Deepfake images generated via InsightFace.]{\begin{centering}
\includegraphics[width=0.98\linewidth]{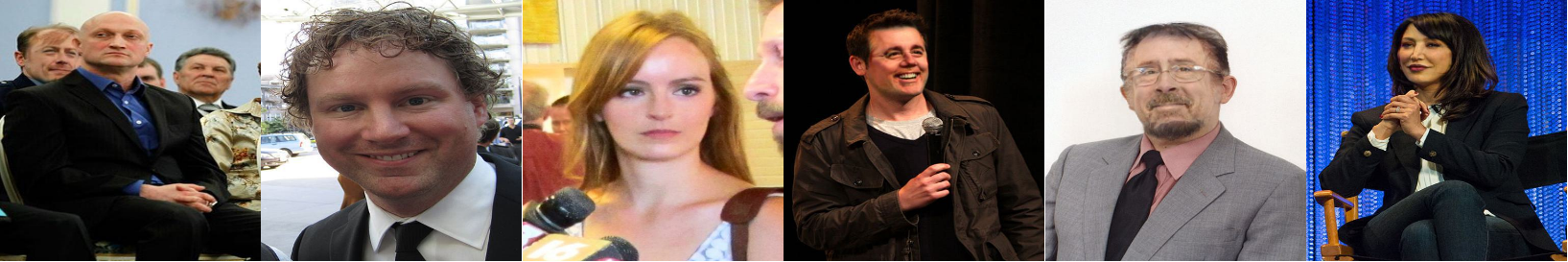}
\end{centering}
}

\end{center}

\caption{The first row contains real image examples from the IMDB-WIKI dataset, the second row contains image examples from FaceForensics++, and the third to fifth rows show image examples generated through our methods.}
\label{fig:deepfake_example}
\end{figure}

In our research, we dig deep into the deepfake phenomenon, understanding its creation and detection~\cite{rossler2019faceforensics++,cao2022end}. Central to our work is a new dataset we've developed, which includes computer-generated images of celebrities. But these aren't ordinary images. They're made using top-tier diffusion methods and a toolset named InsightFace~\cite{insightface}. By capturing a wide range of deepfake techniques, our dataset becomes a valuable tool for building better detection methods. We've shared this dataset online for other researchers, hoping it can drive innovation in deepfake detection. Moreover, we introduce two new ways to test how good these detection tools are. One checks if a tool, trained on a certain type of deepfake, can spot other kinds. The other tests if the tool can still work well on unclear or low-quality images, because real-life photos aren't always perfect.

As we move forward in the paper, we'll detail our approach, discuss what we've learned, and consider the bigger picture of our findings. We'll also touch upon future areas of study crucial for taking on the challenge posed by deepfakes.
\section{Related Works}
This section offers an overview of prevailing deepfake detection datasets and methodologies. 

\subsection{Deepfake Generation}
Recent advancements in synthetic generation and manipulation techniques have underscored the security implications of deepfakes, especially if they are misused for malicious bioinformatics purposes~\cite{arjovsky2017wasserstein,karras2019style,goodfellow2020generative}. Numerous datasets have been crafted to assess the efficacy of detection methods.
The Faceforensics++ dataset~\cite{rossler2019faceforensics++} stands out as a benchmark for evaluating detection capabilities. It employs four manipulation techniques: Deepfakes, FaceSwap, Face2Face, and NeuralTextures. While the first two exchange identities between source and target images, the latter two modify facial expressions while preserving identity. Furthermore, the dataset offers three compression levels—raw, c23 (high quality), and c40 (low quality)—to gauge detectors under varied compression scenarios. 
Celeb-DF~\cite{li2020celeb} focuses on achieving superior visual quality and comprises 590 real videos alongside 5639 synthesized celebrity videos. To address the challenges of detecting manipulated faces in online content, the WildDeepfake dataset~\cite{zi2020wilddeepfake} amasses sequences from deepfake videos found on the internet.
Another noteworthy dataset is the DeepFake Detection Challenge Dataset (DFDC)~\cite{dolhansky2020deepfake}, released by Google. Encompassing over 100,000 clips, DFDC employs a mix of Deepfake, GAN-based, and non-learned methods for synthesis. Nevertheless, many of these datasets involve face-swapping techniques that might leave conspicuous boundaries. The GenImage dataset~\cite{zhu2023genimage}, on the other hand, leverages cutting-edge generation technologies like diffusion models, eliminating telltale stitching signs. Our work, while aligned with GenImage in some aspects, distinguishes itself by specifically targeting synthetic faces of celebrities. Additionally, we introduce novel evaluation tasks, namely cross-generator image classification and degraded image classification, further enriching the evaluation of detection algorithm performance.

\begin{table*}[t]
    \caption{Comparison of fake image detection datasets.}
\begin{center}
        \begin{tabular}{cccccc}
            
            \toprule
             Dataset   & Public & Diffusion & Image Content & Real images & Fake images\\
            \midrule
            UADFV~\cite{yang2019exposing}  & $\usym{2717}$ & $\usym{2717}$ &Face & 241& 252 \\
            FakeSpotter~\cite{wang2019fakespotter} &  $\usym{2717}$ & $\usym{2717}$ & Face & 6,000 & 5,000 \\
            DFFD~\cite{dang2020detection} & $\usym{2713}$ & $\usym{2717}$ & Face & 58,703 & 240,336 \\
            APFFD~\cite{gandhi2020adversarial} & $\usym{2713}$ & $\usym{2717}$ & Face & 5,000 & 5,000 \\
            ForgeryNet~\cite{he2021forgerynet} & $\usym{2713}$ & $\usym{2717}$ & Face & 1,438,201 & 1,457,861 \\
            GenImage\cite{zhu2023genimage}  & $\usym{2713}$ & $\usym{2713}$ & General  & 1,331,167& 1,350,000 \\
            DeepFakeFace(Ours)  & $\usym{2713}$ & $\usym{2713}$  &Face  &  30,000 & 90,000 \\    
            \bottomrule
        \end{tabular}
    \end{center}
    \label{table:dataset_compare}
\end{table*}

\subsection{Deepfake Detection}
Spatial-based detection techniques have gained prominence in recent times. Notably, the Face x-ray method~\cite{li2020face} discerns deepfakes by predicting the presence of blending boundaries. Similarly, Shiohara \textit{et al.}~\cite{shiohara2022detecting} employ image blends of the same identity. 
Zhu~\cite{zhu2021face} integrated 3D decomposition into detection, devising the FD2Net, which synergizes input images with extracted facial details. Identity Consistency Transformer (ICT)~\cite{dong2022protecting} harnesses publicly available videos, gauging the consistency between inner and outer face regions. Viewing detection as a fine-grained classification challenge, Zhao \textit{et al.}~\cite{zhao2021multi} developed a multi-attentional architecture network to capture local features. 
Many methods also harness frequency clues for detection~\cite{qian2020thinking,liu2021spatial,luo2021generalizing,li2021frequency}. Qian \textit{et al.}~\cite{qian2020thinking}, for instance, harness frequency-aware decomposition and local frequency statistics, while Spatial-Phase Shallow Learning~\cite{liu2021spatial} exploits both spatial images and phase spectra. Additionally, the exploration of auditory modalities and temporal information has paved the way for innovative detection strategies~\cite{zhou2021joint,haliassos2021lips,masi2020two,sun2021improving,zheng2021exploring,cozzolino2021id}.
In assessing the potency of our DFF dataset, we adopted the state-of-the-art RECCE method~\cite{cao2022end}, which demonstrated exceptional prowess in both cross-generator image classification and degraded image classification.
\section{Methodology}
\label{section:method}
In this section, we detail the construction and characteristics of our DeepFakeFace (DFF) dataset, the generative models employed for synthesizing deepfakes, and the comprehensive process underlying fake image generation.

\subsection{Dataset Details}

We present a new dataset named DeepFakeFace (DFF) to assess the ability of deepfake detectors to distinguish AI-generated and authentic images. There are 30,000 pairs of real and fake images. Since we aim to protect the privacy of celebrities, 30,000 real images of dataset all comes from IMDB-WIKI dataset~\cite{rothe2015dex}. The dataset consists of 120,000 images which incorporate 30,000 real images and 90,000 fake images.  We employ three different generative models for synthesizing deepfakes: Stable Diffusion v1.5, Stable Diffusion Inpainting and a powerful toolbox InsightFace. Each model generates 30,000 fake images.

\subsection{Fake Image Generators}

Diffusion models~\cite{ho2020denoising}, which create high-resolution images via the sequential deployment of denoising autoencoders, are an integral part of our methodology. Direct pixel-level operation, however, proves resource-intensive in terms of time and computational complexity. To counteract this, stable diffusion~\cite{rombach2022high} harnesses diffusion models within the latent space. This not only conserves computational resources but also maintains the quality and flexibility of generated images. With its prowess in synthesizing photo-realistic images from any given input text, we adopted stable diffusion for crafting deepfakes. These images bear a resolution of 512 × 512. Our study utilizes both the Stable Diffusion v1.5 and Stable Diffusion Inpainting models. Additionally, for a multifaceted approach, the InsightFace~\cite{insightface} toolbox—equipped with top-tier algorithms for face recognition, detection, alignment, and swapping—also contributes to our deepfake generation.

\subsection{Fake Image Generation Process}

The IMDB-WIKI dataset, known for its extensive compilation of face images annotated with gender and age, is our primary source of authentic images. Leveraging this dataset allows for effortless extraction of gender and age metadata from its label files. For consistency, images are randomly matched based on gender, and this configuration is adhered to in subsequent methodologies. 
Upon retrieval of gender, age, and identity for each image, prompts corresponding to each image are generated. These prompts adhere to the template: "name, celebrity, age", where "name" and "age" are replaced by the image's actual identity and age, respectively. Though the IMDB-WIKI dataset furnishes aligned faces with the original facial bounding box, discrepancies in accuracy were noted in some bounding boxes. To address this, we utilized the cutting-edge RetinaFace face detector~\cite{deng2020retinaface} to redefine facial bounding boxes and generate corresponding mask images. 
Equipped with this refined data, deepfakes are then generated using Stable Diffusion v1.5, Stable Diffusion Inpainting, and InsightFace, respectively.
\section{Experiments}
In this section, we look at how well two new tasks perform: classifying images from different generators and classifying degraded images. We use three popular metrics—Accuracy (Acc), Area Under the Receiver Operating Characteristic Curve (AUC), and Equal Error Rate (EER)—to measure deepfake detection performance. A lower EER means the detector is more accurate, while higher Acc and AUC indicate better performance~\cite{cao2022end}.


\subsection{Cross-generator Image Classification}
Table~\ref{table:cross_generator} showcases the variability in RECCE's performance~\cite{cao2022end} across different deepfake generation techniques:

{\bf Stable Diffusion v1.5}: The results indicate that this method is exceptionally challenging for RECCE to handle. With an accuracy of just 38.14\%, RECCE struggles significantly against deepfakes that are constructed entirely from scratch, including both facial and background elements. This poor performance suggests that deepfakes generated in this manner possess features that the RECCE model, when trained on conventional datasets like FaceForensics++~\cite{rossler2019faceforensics++}, struggles to identify correctly.

{\bf Stable Diffusion Inpainting}: Deepfakes generated by viewing the creation process as an inpainting problem result in a slightly better performance by RECCE, with an accuracy of 51.35\%. While this is an improvement from the previous method, it's still an evident challenge, indicating that synthesizing just the facial area (and retaining the background) can still effectively deceive the detector.

{\bf InsightFace}: Among the three methods, RECCE performs the best against deepfakes produced by InsightFace, recording an accuracy of 58.99\%. However, this marginally surpasses random guessing, suggesting that even in the best-case scenario, RECCE finds it strenuous to pinpoint deepfakes generated by this method. The method, which involves swapping identities between source and target images, introduces complexities that even state-of-the-art detectors like RECCE grapple with.

Given these insights, it's evident that the technique utilized for generating deepfakes critically influences a detector's performance. There's an urgent need to evolve current detection models, like RECCE~\cite{cao2022end}, to keep pace with rapidly advancing deepfake generation methods.

\begin{table}[t]
  \caption{Performance of RECCE~\cite{cao2022end} across different generators, measured in terms of Acc (\%), AUC (\%), and EER (\%).}
  \label{table:cross_generator}
\begin{center}
\begin{tabular}{cccc}
\toprule
Method & Acc & AUC & EER\\
\hline
Stable Diffusion v1.5 & 0.3814 & 0.3512 & 0.6188\\
Stable Diffusion Inpainting & 0.5135 & 0.5152 & 0.4961\\
Insight & 0.5899 & 0.6312 & 0.4105\\
\bottomrule
\end{tabular}
\end{center}

\end{table}

\begin{table}[b]
  \caption{Robustness evaluation in terms of ACC(\%), AUC (\%) and EER(\%).}
  \label{table:table_robustness}
\begin{center}
\begin{tabular}{ccccc}
\toprule
Perturbation & Method & Acc & AUC & EER \\
\hline
                 
                 &  Stable Diffusion v1.5        & 0.3901  & 0.3637  & 0.6106 \\ 
Color Contrast   &  Stable Diffusion Inpainting  & 0.5141  & 0.5139  & 0.4975\\
                 &  Insight                      & 0.5815  & 0.6170  & 0.4213   \\
                 \hline
                 
                 &  Stable Diffusion v1.5        & 0.3921  & 0.3607  & 0.6079 \\ 
Color Saturation &  Stable Diffusion Inpainting  & 0.5226  & 0.5357  & 0.4797\\
                 &  Insight                      & 0.5824  & 0.6193  & 0.4200   \\
                 \hline
                 
                 &  Stable Diffusion v1.5        & 0.5268  & 0.5905  & 0.4223 \\ 
Gaussian Blur    &  Stable Diffusion Inpainting  & 0.5050  & 0.4878  & 0.5080\\
                 &  Insight                      & 0.5111  & 0.5061  & 0.4977   \\
                 \hline
                 
                 &  Stable Diffusion v1.5        & 0.4249  & 0.3928  & 0.5782 \\ 
Pixelation       &  Stable Diffusion Inpainting  & 0.5037  & 0.4898  & 0.5069\\
                 &  Insight                      & 0.5459  & 0.5825  & 0.4469   \\
\bottomrule
\end{tabular}
\end{center}
\end{table}

\subsection{Degraded Image Classification}

In the digital age, images are frequently subjected to various alterations when shared across online platforms. Consequently, it's crucial for deepfake detectors to maintain their performance despite these degradations. Based on prior research~\cite{jiang2020deeperforensics,haliassos2021lips,cao2022end}, we evaluated the robustness of RECCE to common image perturbations, encompassed under Image Quality Assessment (IQA)~\cite{ponomarenko2015image}. These perturbations include changes in color saturation, color contrast, Gaussian blur, and pixelation.

Comparing the results from Table \ref{table:cross_generator} and Table \ref{table:table_robustness} provides some intriguing insights:

{\bf Color Contrast}: The effects of color contrast adjustments seem to have minimal impact on the detection capabilities of RECCE across the tested deepfake generation methods. In fact, there are only slight variations in the accuracy for the tested methods when compared to their baseline performance, with Stable Diffusion v1.5 recording a negligible increase.

{\bf Color Saturation}: Color saturation changes also exhibit minor deviations from the baseline results. Notably, the Stable Diffusion Inpainting method sees an increase in accuracy, implying that alterations in saturation might slightly benefit RECCE's detection capabilities for this specific generation technique.

{\bf Gaussian Blur}: Interestingly, the Gaussian blur appears to have a positive influence on RECCE's performance, especially for the Stable Diffusion v1.5 method, which sees a substantial boost in accuracy. This may hint at certain inherent characteristics of these deepfakes becoming more apparent or detectable with blur.

{\bf Pixelation}: Pixelation introduces some degradation in RECCE's performance for the Stable Diffusion v1.5 method, but the other two generation techniques still show competitive results, especially when compared to their baseline metrics.

In summary, while we initially hypothesized that image perturbations would pose significant challenges for the RECCE model, the data suggests otherwise. Instead of detrimental impacts, some perturbations seem to inadvertently assist RECCE in detecting certain deepfakes. This observation emphasizes the complexity of the deepfake detection landscape and calls for a deeper exploration into how different image alterations interact with detection algorithms.

\section{Conclusions}
\label{sec:conlusion}

In this investigation, we immersed ourselves into the intricate realm of deepfake image generation and detection, placing particular emphasis on the robustness and adaptability of detection algorithms. At the heart of our research lies the DeepFakeFace (DFF) dataset, a rich collection of synthetic celebrity images created using diffusion models. We introduced two innovative evaluation tasks—cross-generator image classification and degraded image classification—to gauge the versatility of deepfake detectors. Our findings underline the varied detectability across different generative techniques and diverse image conditions, shedding light on the pressing need for evolving detection strategies and a granular exploration into how various generative models and image perturbations impact detection efficacy.
Significantly, we hold high expectations for our open-sourced DeepFakeFace (DFF) dataset to serve as a potent tool in the fight against deepfakes. We earnestly encourage the wider community to leverage this resource, hoping that it catalyzes the emergence of more efficient and universally applicable deepfake detection strategies. In conclusion, our endeavor provides a pivotal contribution to the field of deepfake detection, offering invaluable insights and tools poised to stimulate further research and advancements in this domain.


\bibliography{references.bib}
\bibliographystyle{plain}




\end{document}